\documentclass{article}
\usepackage{ijcai19}

\usepackage{times}
\usepackage{xcolor}
\usepackage{soul}
\usepackage[utf8]{inputenc}
\usepackage[small]{caption}
\usepackage{url}
\usepackage[hidelinks]{hyperref}
\usepackage{booktabs}

\usepackage{amsmath}
\usepackage{amssymb}
\usepackage{bm}
\usepackage{graphicx}

\newcommand{\RR}{\mathbb{R}}
\newcommand{\PP}{\mathbb{P}}
\newcommand{\EE}{\mathbb{E}}

\newcommand{\cA}{\mathcal{A}}
\newcommand{\cI}{\mathcal{I}}
\newcommand{\cY}{\mathcal{Y}}

\newcommand{\ca}{\boldsymbol{a}}

\title{Recurrent Existence Determination Through Policy Optimization}

\author{
Baoxiang Wang
\affiliations
The Chinese University of Hong Kong \\
\emails
bxwang@cse.cuhk.edu.hk
}

\begin{document}

\maketitle

\begin{abstract}
Binary determination of the presence of objects is one of the problems where humans perform extraordinarily better than computer vision systems, in terms of both speed and preciseness. One of the possible reasons is that humans can skip most of the clutter and attend only on salient regions. Recurrent attention models (RAM) are the first computational models to imitate the way humans process images via the REINFORCE algorithm. Despite that RAM is originally designed for image recognition, we extend it and present recurrent existence determination, an attention-based mechanism to solve the existence determination. Our algorithm employs a novel $k$-maximum aggregation layer and a new reward mechanism to address the issue of delayed rewards, which would have caused the instability of the training process. The experimental analysis demonstrates significant efficiency and accuracy improvement over existing approaches, on both synthetic and real-world datasets.
\end{abstract}

\section{Introduction}

Object existence determination\footnote{Certain literature \cite{zagoruyko2016multipath} may refer the problem using the terminology \textit{object detection}. More commonly object detection refers to both deciding the existence of certain patterns and subsequently locating them if so.} 
(ED) focuses on deciding if certain visual patterns exist in an image.
As the basis of many computing vision tasks, ED's quality affects further processing such as locating certain patterns (apart from telling the existence), segmentation of certain patterns, object recognition, and object tracking in consecutive image frames.
However, while ED is conducted by humans rapidly and effortlessly \cite{das2016human,borji2014salient}, the performance of computer vision algorithms are surprisingly poor especially when the image is of large size and low quality.
Hence, it is desirable to develop efficient and noise-proof systems to deal with object detection tasks with large and noisy images.

In fact, the way humans process images is not similar to recent prevailing approaches such as detecting objects via convolution networks (ConvNet) and residual networks \cite{he2016deep}.
Instead of taking all pixels from the image in parallel, humans perform sequential interactions with the image. 
Humans may recursively deploy visual attentions and performs glimpses on selective locations to acquire information.
At the end of the processing, information from all past locations and glimpses is gathered together to make the final decision. 
Such behavior accomplishes ED tasks efficiently, especially for large images as it depends only on the number of saccades.
Meanwhile, as the approach selectively learns to skip the clutter\footnote{Clutter are the irrelevant features of the visual environment, discussed in RAM \cite{mnih2014recurrent}.}, it tends to be less sensitive to noise compared with those that take all pixels into the computation.

The process can be naturally interpreted as a reinforcement learning (RL) task where each image represents an environment.
At the beginning of the process, the agent conducts an action which is represented by a 2-dimension Cartesian coordinate.
When the environment receives the action, it calculates the retina-like representation of the image at the corresponding location, and returns that representation to the agent as the agent's observation.
Repeatedly until the last step, the agent predicts the detection result based on the trajectory and receives the evaluation of its prediction as the reward signal.
It is important to note that the agent has never had access to the full image directly.
Instead, it carefully chooses its actions in order to get the desired partial observations of the internal states of the environment.

Recurrent attention models (RAM) \cite{mnih2014recurrent} are the first computational models to imitate the process with a reinforcement learning algorithm.
The success of RAM leads to enormous studies on attention-based computer vision solutions \cite{yeung2016end,gregor2015draw}.
However, RAM and their extensions \cite{ba2015learning,ba2014multiple} are designed to solve object recognition tasks such as handwritten digit classification.
Those models largely ignore the trajectory information which causes massively delayed rewards.
Indeed in RAM, the reward function is associated with only the last step of the process and is otherwise zero.
The actions before that, which deploy the attention for the model, do not receive direct feedback and are therefore not efficiently learned.
Especially in ED (and in general, object detection) tasks, delayed rewards fail to provide reinforcement signals to the choice of locations when the glimpse at certain locations may provide explicit information for the existence of the object.

We present recurrent existence determination models (RED), which inherit the advantage of RAM that the attention is only deployed on locations that are deemed informative.
Our approach involves a new observation setting which allows the agent to have access to explicit visual patches.
Unlike previous trails which blur the pixels that are far from the saccade location, we acquire the exact patches which help to detect the existence of specific patterns.
We employ gated recurrent units (GRU) \cite{chung2014empirical} to encode the historical information acquired by the agent and generates temporary predictions at each time step.
The temporary predictions over the time horizon are then aggregated via a novel $k$-maximum aggregation layer, which averages the $k$-greatest value to compute the final decision.
It allows the rewards to be backpropagated to the early and middle stages of the processing directly apart from through the recurrent connections of the GRU.
It provides immediate feedback which guides the agent to allocate its attention, and therefore addresses the issues caused by delayed rewards.

RED is evaluated empirically on both synthetic datasets, \textit{Stained} \textit{MNIST}, and real-world datasets.
Stained MNIST is a set of handwritten digits from MNIST. Additionally, the resolution has been enlarged and each digit may be added dot stains around the writings.
The dataset is designed to compare the performance of RED and existing algorithms on images with high-resolution settings.
The results show that attention based models run extraordinarily faster than traditional, ConvNet-based methods \cite{dieleman2015rotation,graham2014fractional}, while having better accuracy as well.
Experiments on real-world dataset show superior speed improvement and competitive accuracy on retinopathy screening, compared to existing approaches.
This also demonstrates that our algorithm is practical enough to be applied to real-world systems.

\section{Preliminaries}
\label{sec:prelim}

\subsection{Policy Gradients}
\label{sec:prelim-rl}

In an episode of RL \cite{sutton2018reinforcement}, at each time step $t$, the agent takes an action $a_t$ from the set $\cA_t$ of feasible actions.
Receiving the action from the agent, the environment updates its internal state, and returns an observation $x_t$ and a scalar reward $r_t$ to the agent accordingly.
In most of the problems, the observation does not fully describe the internal state of the environment, and the agent has to develop its policy using only the partial observations of the state.
This process continues until the time horizon $T$.
Let $R_t=\sum_{t^\prime=1}^{t^\prime=t}r_{t^\prime}$ denote the cumulative rewards up to time $t$, the policy is trained to maximize the expectation of $\EE[R_T]$.
Let $\pi_\theta$ be the policy function, parameterized by $\theta$, the REINFORCE algorithm \cite{williams1992simple,mnih2016asynchronous} estimates the policy gradient using
\begin{equation}
g = \EE_\pi[\nabla_\theta\log\pi(a_t|s_t)(R_t-b_t)],
\end{equation}
where $b_t$ is an baseline function for variance reduction.

It is common in RL to use $x_t$ as the state $s_t$. 
However, consider that $x_t$ are small patches in our setting, the information in a single $x_t$ is insufficient.
Ideally, the decision of the action is based on the trajectory $\tau_t=(a_1,x_1,r_1,\dots,a_{t-1},x_{t-1},r_{t-1})$ which includes past actions, observations, and rewards.
To handle the growing dimensionality of $\tau_t$, the agent maintains an internal state\footnote{In our paper, $s_t$ is defined to be the state of the agent instead of the state of the environment.} $s_t$ which encodes the trajectory using a recurrent neural network (RNN), and updates it repeatedly until the end of the time horizon.
In this way, the action is decided by the policy function, based only on the internal state $s_t$ of the agent.
Note that the full state of the environment is $\tau_t$ and the image to be processed, and the agent observes $\tau_t$ only.

The training of RL repeats the above process from step $1$ to step $T$ for a certain number of episodes.
At the beginning of each episode, the agent resets its internal state while the environment resets its internal state as well.
The model parameters are maintained across multiple episodes and are updated gradually as the occurrence of the reward signals at the end of each episode.
Note that in RAM and RED, different from general online learning framework, the agent does not receive the reward signal in the middle stages of an episode.
Hence the reward signals are inevitably heavily delayed, and RED need to address temporal credit assignment problem \cite{sutton1984temporal}, which evaluates individual action within a sequence of actions according to a single reinforcement signal.

\subsection{Glimpse and Retina-Like Representations} \label{sec:attention}

A retina-like representation is the visual signal humans receive when glimpsing at a point of an image.
The visual effect is that regions close to the focused location tend to retain their original, high-resolution form, while regions far from the focused location are blurred and passed to the human brain in their low-resolution form.
In RAM and RED, the environment calculates the retina-like representations and returns it as the observation.
Existing approaches to mimic such visual effects have been used in RAM and RAM's variants.
They can be categorized into two classes: soft attention \cite{gregor2015draw,xu2015show} and hard attention \cite{eslami2016attend,xu2015show,mnih2014recurrent}. 

Soft attention \cite{hermann2015teaching} applies a filter centered at the focused location.
It imitates human behaviors, which downsamples the image gradually as it approaches far away from the focused point, resulting in a smooth representation.
The approach is fully differentiable and is hence amenable to be trained straightforwardly using neural networks together with gradient descent.
Despite those merits, soft attention is in general computationally expensive as it involves the filtering operation over all pixels, which deviates from the idea of RED and RAM to only examines parts of the image and subsequently making the process relatively inefficient.

Hard attention, on the other hand, extracts pixels with predefined sample rates.
Fewer pixels are extracted as the region approaches further away from the focused location, making the process cost only constant time.
Hard attention fits the idea of RED well though it is non-differentiable as it indexes the image and extracts pixels.
To address the non-differentiability, we develop our training algorithm via policy gradient and use the rollouts of the attention mechanism to estimate the gradient.
Formally, let $x_t$ be a list of $c$ channels and the $i$-th channel extracts the squared region centered at $a_t$ with size $n_i\times n_i$, and down sample the patch to $n_1\times n_1$.
The channels are incorporated with the location information (known as the what and where pathways) with the patch information by adding the linear transformation $\tanh(W_{xa}a_t)$ of $a_t$.
Note that the value of each entry $W_{xa}$ will be restricted to be relatively small compared to the pixel values to retain the original patch information.

\subsection{Convolutional Gated Recurrent Units}

RAM and RED use an RNN to encode the trajectory and update the states of the agent.
While both long short-term memory (LSTM) and GRU are prevailing RNN implementations in sequential data processing \cite{chung2014empirical}, GRU is preferred to LSTM in RED.
The reason is that the input passes through the unit explicitly when a GRU merges the memory cell state and the output state in an LSTM unit.
This explicit information helps to make the temporary detection decisions and enables our design of the $k$-maximum aggregation layer.
Meanwhile, with the merge, the agent updates its internal state more efficiently.
The speed improvement is critical especially for real-time applications such as surveillance anomaly detection when an instant detection decision is required.

\begin{figure}[t]
\includegraphics[width=0.48\textwidth,height=0.26\textwidth]{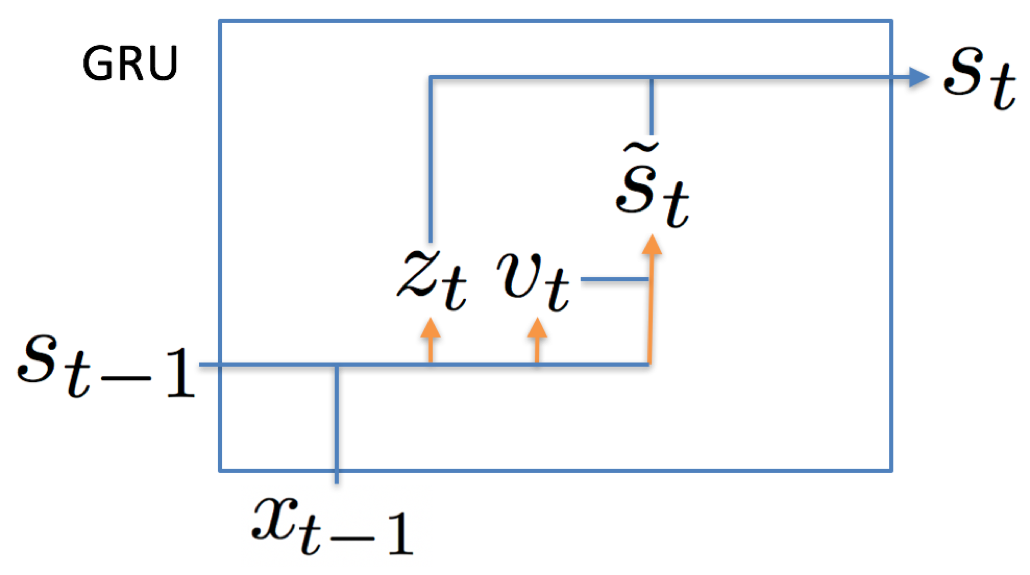}
\caption{Illustration of convolutional gated recurrent units.}
\label{fig:gru}
\end{figure}

We use convolutional GRU, a variant of GRU where the matrix product operations between the output state $s_t$, the input $x_t$, and the model parameters are replaced with convolution operations, and $s_t$ and $x_t$ are kept in their 2-dimension matrix shapes \cite{xingjian2015convolutional}.
A graphical illustration of GRU is shown in Figure \ref{fig:gru}, where lines in orange represent convolution operation.
The gate mechanism in a convolutional GRU is formulated as 
\begin{align}
\begin{split}
z_t & = \sigma(W_{zh}\ast s_{t-1}+W_{zx}\ast x_t) \\
v_t & = \sigma(W_{rh}\ast s_{t-1}+W_{rx}\ast x_t) \\
\tilde{s}_t & = \tanh(W_{sh}\ast (v_t\circ s_{t-1})+W_{sx}\ast x_t) \\
s_t & = (1-z_t)s_{t-1} + z_t\tilde{s}_t, \label{eqn:gru}
\end{split}
\end{align}
where $\ast$ denotes convolution, $\circ$ denotes Hadamard product, $\sigma(\cdot)$ denotes the sigmoid function, and $z_t$ and $v_t$ are the update gate and the reset gate, respectively.
$W_{zh}$, $W_{zx}$, $W_{rh}$, $W_{rx}$, $W_{sh}$, and $W_{sx}$ are trainable parameters.
Convolutional GRU retains the spatial information in the output state so that temporary detection decisions can be made well before the end of an episode.

\section{Recurrent Existence Determination}

\begin{figure}[t]
\includegraphics[width=0.48\textwidth,height=0.26\textwidth]{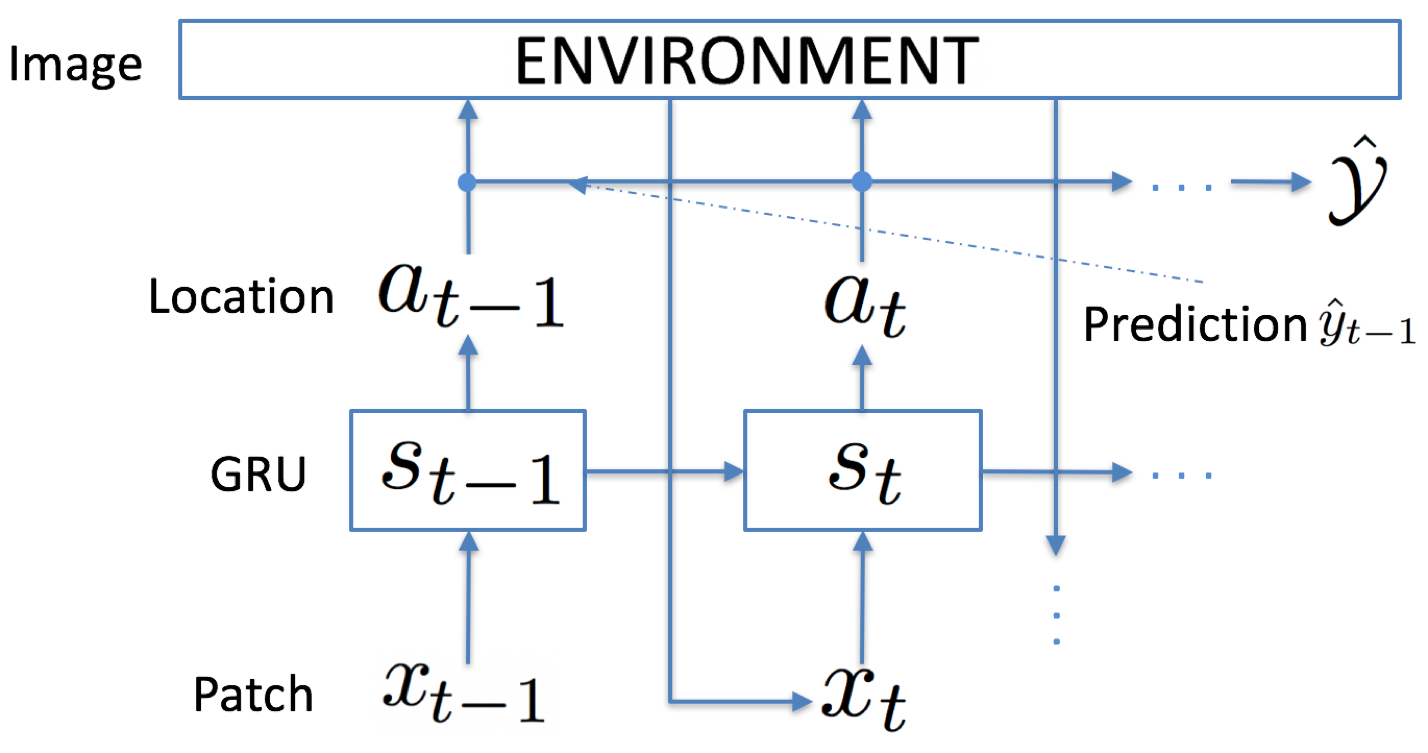}
\caption{The attention mechanism and the reward mechanism in our proposed RED model.}
\label{fig:arch}
\end{figure}

In this section we discuss the three main components of RED, namely, the attention mechanism, the $k$-maximum aggregation layer, and the policy gradient estimator.
Taking together, an illustrative of our model is shown in Figure~\ref{fig:arch}, where the arrows denote forward propagation.

\subsection{Attention Mechanism in RED}

We formulate the attention mechanism within each of the episodes, that is, within the processing of one image.
Let $\cI$ denote the image, the agent has $x_0$, which is the low-resolution form of $\cI$, as the initial observation.
The state $s_0$ of the agent is initialized as the zero vector.
Repeatedly, at each time step $t$, the agent calculates its action $a_t\in \RR^2$, according to
\begin{eqnarray}
a_t & = & \tanh(W_{as}s_t) + \epsilon_t, \label{eqn:att}
\end{eqnarray}
where $W_{as}$ is a trainable parameter of the model and $\epsilon_t$ is a random noise to improve exploration.
The action $a_t$ refers to a Cartesian Coordinate on the image, with $(-1,-1)$ corresponding to the bottom-left corner of $\cI$ and $(1,1)$ corresponding to the top-right corner of $\cI$.
Each entry of $\epsilon_t$ is sampled from a normal distribution with a fixed standard deviation of $\beta$, independently.
The environment returns the retina-like representation $x_t$ via the hard attention model described in Section~\ref{sec:attention}.

The agent employs a single convolutional GRU and uses the output states $s_t$ as the agent's state, defined in Equation~\eqref{eqn:gru}.
The state $s_t$ has the same shape $n_1\times n_1$ as each channel of the observation, which is ensured by the convolution operation in Equation~\eqref{eqn:gru}.
We take advantage of GRU that the reset gate $v_t$ in Equation~\eqref{eqn:gru} controls the choice between long-term dependencies and short-term observations.
The former is important for exploring future attention deployment within an episode, while the later is important for exploiting currently available information to make temporary decisions.
By training over a large number of episodes, the agent learns to balance exploration and exploitation from the reinforcement signals by updating its gate parameters.

With Equation~\eqref{eqn:gru}, Equation~\eqref{eqn:att} and the hard attention mechanism, each rollout is computed in constant time with respect to the image size as $T$ and $n_i$ are fixed. As a result, a trained RED model is able to make predictions very efficiently.

\subsection{Prediction Aggregation}

We present the framework to generate temporary predictions and subsequently aggregate the temporary predictions into the final prediction, i.e. the detection of the patterns.
At each time step $t$, the agent has access to the output state $s_t$ of the GRU which carries the information from the current patch $x_t$.
Based on $s_t$ the agent makes a temporary prediction $\hat{\cY}$ using a feed-forward network followed by a non-linear operation $\hat{y}_t=\frac{1}{2}(1+\tanh(W_{ys}s_t))$,
where $\hat{y}_t$ is the estimated probability that the object exists in $\cI$.

The temporary predictions are aggregated over time using our newly proposed $k$-maximum aggregation layer.
The layer calculates the weighted average of the top $k$ largest values among $\hat{y}_{t_0},\cdots ,\hat{y}_T$, where $t_0\geq 1$ is a fixed threshold of the model.
The output $\hat{\cY}$ of the $k$-maximum layer is formulated as
\begin{equation}
\hat{\cY}=\frac{1}{Z}\sum_{t\in K}(1-\gamma^t) \hat{y}_t, \label{eqn:kmax}
\end{equation}
where $K=k\text{-argmax}_{t_0\leq t \leq T} \{\hat{y}_t\}$ is the set of the indexes of the top $k$-largest temporary predicted probabilities, and $Z=\sum_{t\in K}(1-\gamma^t)$ is the normalizer to guarantee $0\leq \hat{\cY} \leq 1$.
In Equation~\eqref{eqn:kmax} we elaborate a time discount factor $1-\gamma^t$ which assigns a larger value toward the late stages of the process than the early stages of the process, where $\gamma$ is fixed through the process.
The factor $\gamma$ is a trade-off between RAM where all previous steps are used to benefit the prediction at the end of the episode and majority voting where all observation contributes to the binary determination.  

The advantage of using the $k$-maximum layer is to guide the model to balance between exploration and exploitation\footnote{It also helps to address the problem of vanishing gradient at the same time, though, it is out of the scope of this paper.}.
Consider that only steps $t$ with top $k$ largest $\hat{y}_t$ are taken into account in the final prediction, the model has a sufficient number of time steps to explore different locations on $\cI$ and does not need to worry about affecting the final prediction.
In fact, exploring the context of the image is important to collect information and locate the detection objective in late stages.
The time discount factor further reinforces that by assigning larger weights toward late stages, which encourages the agent to explore at the early stages of the process and exploit at the late stages of the process.

Viewing our proposed prediction aggregation mechanism from an RL perspective, it addresses the credit assignment problem \cite{sutton1984temporal}.
Existing studies on applications via policy learning, e.g. \cite{mnih2016asynchronous,li2018policy,young2018metatrace}, commonly equally assign the feedback of an episode toward all actions the agent has made.
The large variance of estimating the quality of a single action using the outcome of the entire episode is neutralized by training the agent for millions of episodes.
However, in our settings the state of the environment is diverse as each different image $\cI$ corresponds to a unique initial state of the environment.
The variance cannot be reduced by simply training on a large dataset of images without a fixed observation function with respect to $\cI$.
In this way, our proposed aggregation mechanism is necessary to help the algorithm to converge and it is the key component for RED to make detection decisions.

\subsection{Policy Gradient Estimation}

In this section we derive the estimator of the policy gradient. It is feasible to apply the policy gradient theorem \cite{sutton2000policy,sutton2018reinforcement,mnih2014recurrent}, but since we know the exact formulation of the reward function we can largely reduce the variance by incorporating this information. 
To achieve this, we derive the estimator specifically for RED from scratch by taking the derivative of the expected cumulative regret, defined as the negative reward \cite{li2016contextual}. 

Let $W$ denote the set of trainable parameters including $\theta$, $W_{as}$, $W_{xa}$, $W_{ys}$ and the trainable parameters in the GRU, in Equation~\eqref{eqn:gru}.
Also let $\cY\in \{0,1\}$ be the ground truth of the detection result, where $0$ and $1$ correspond to the existence and non-existence of the object, respectively.
Define a rollout $\hat{\tau_T}$ of the trajectory within an episode to be a sample drawn from the probability distribution $\PP(\tau_T|\pi_\theta(\cdot))$.
During training, the agent generates its rollouts $\hat{\tau}_T$ and predictions $\hat{\cY}$ on an iterator of $(\cI, \cY)$ pairs, where each pair of the image and the ground truth corresponds to one episode of RL.
Define the regret $L_T$ to be the squared error between the predicted probability and the ground truth
\begin{equation}
L_T=(\hat{\cY}-\cY)^2.
\end{equation}
The model updates $W$ after the conclusion of each episode, when it receives a reward signal $r_T=1-L_T$.
In this case, $L_T+R_T=L_T+r_T=1$.

We utilize similar arguments in the policy gradient theorem to address the non-differentiability.
Let $\ca=(\hat{a}_1,\dots,\hat{a}_T)$ be the sequence of actions in $\hat{\tau}_T$, we have the expected regret
\begin{equation}
\EE[L_T|W]=\sum\nolimits_{\ca}\PP(\ca|W)(\hat{\cY}_{\ca}-\cY)^2,
\end{equation}
where the deterministic variable $\hat{\cY}_{\ca}$ denotes the model's counterfactual prediction under the condition that $\ca$ is sampled with probability one.
Since there is no randomness involved on the environment side, the expectation above is calculated over the actions only.
Taking the derivative with respect to $W$, the gradient is
\begin{align}
\begin{split}
\nabla_{W}\EE[L_T|&W]  = \EE_{\ca\sim \PP(\ca|W)}[(\hat{\cY}_{\ca}-\cY)^2 \\
& \nabla_{W}\log\PP(\ca|W) + 2(\hat{\cY}_{\ca}-\cY)\nabla_{W}\hat{\cY}_{\ca}], \label{eqn:policy}
\end{split}
\end{align}
where the immediate partial derivative from the chain rule of Equation~\eqref{eqn:att} is
\begin{equation}
\nabla_{W}\log\PP(a_t|W)=\frac{1}{\beta^2}\cdot(a_t-\EE[a_t|W])s_{t-1}^T. \label{eqn:log-action}
\end{equation}
Further, deduct from the regret the baseline function
\begin{equation}
b_T=\EE_{\ca\sim \PP(\ca|W)}[(\hat{\cY}_{\ca}-\cY)^2],
\end{equation}
which calculates the expected regret from the rollouts, for variance reduction. 
By doing this we account only the difference between the actual reward and the baseline function.
Note that the baseline function introduces no bias into the expectation in Equation~\eqref{eqn:policy}, while it is used to reduces the variance when estimating the policy gradient using the Monte-Carlo samples $\ca$.
At the end of each episode, update $W$ according to
\begin{align}
\begin{split}
W  \leftarrow\text{ } & W - \alpha\EE_{\ca\sim \PP(\ca|W)}[((\hat{\cY}_{\ca}-\cY)^2-b_T) \\
& \nabla_{W}\log\PP(\ca|W) + 2(\hat{\cY}_{\ca}-\cY)\nabla_{W} \hat{\cY}_{\ca}], \label{eqn:bias}
\end{split}
\end{align}
where $\alpha$ is the learning rate.

To estimate the expectation in Equation~\eqref{eqn:bias}, the agent generates a rollout $\hat{\tau}_T$ which samples $\ca$ according to $\ca\sim \PP(\ca|W)$.
The expectation is then estimated using the generated $\ca$ value by Equation~\eqref{eqn:log-action} and REINFORCE's back-propagation \cite{wierstra2007solving}.
Note that the second part $2(\hat{\cY}_{\ca}-\cY)\nabla_{W}\hat{\cY}_{\ca}$ of the gradient is useful, though, it is sometimes ignored in previous studies \cite{mnih2014recurrent}. 
It connects the regret to the early stages which allows the regret signal to be back-propagated directly to those steps and to guide the exploitation of the agent.
It can be regarded as a retrospective assignment of the credits after the rollout has been fully generated, equivalently making the reward $r_t$ in RED no longer $0$ when $t<T$ during the training phase, which addresses the issues caused by delayed rewards.

\section{Experiments}

\subsection{Stained MNIST}

We first test and compare RED on our synthetic dataset, \textit{stained MNIST}, with a variety of baseline methods.
Stained MNIST contains a set of handwritten digits, which have very high resolution and much thinner writings than the original MNIST does.
Each digit may be associated with multiple \textit{stains} on the edge of its writing, which are dot-shaped regions with high tonal value.
The algorithms are required to predict if such stains exist in the images. 
The task is very challenging as the image resolution is very high while the writings are thin and unclear.
Hence it is hard to locate the stains or recognize the stains from the writings.

Stained MNIST is constructed by modifying the original MNIST dataset as follows.
Each image from MNIST is first resized to $7168\times 7168$ by bilinear interpolation, and rescaled to 0 to 1 tonal value. 
The enlarged images are then smoothed using a Gaussian filter with a $20\times 20$ kernel.
After that, it calculates the central differences of each pixel and finds out the set $C$ of pixels with $0.2$ or larger gradient. 
The tonal values of pixels that are within $500$ pixels of $C$ are set to $0$.
This operation makes the writings of the digits much thinner in the high-resolution images.
After removing those pixels, the gradient of each pixel is calculated again, and $10$ to $15$ stains with radius $12$ are randomly added at pixels with high gradient.

The hyper-parameters of RED are set to be $c=3,n_1=18,n_2=36,n_3=54$ for attention mechanism and $\gamma=0.95, k=25, t_0=10$ for prediction aggregation, through a random search on a training subset. 
The search over $\gamma\in [0.9, 0.98]$, $k\in [15,30]$, and $t_0\in [10,50]$ does not observe significant difference on the performance.
Accordingly, the patch size $x_t$ is set to $n_1\times n_1$ as the input of the GRU.
The horizon is fixed to $T=350$, where no significant improvement can be observed by further increasing it.
When estimating the baseline function $b_T$, 15 instances are sampled and are averaged over.
When evaluating RED, we remove the stochastic components $\epsilon_t$ in computing the actions.

We compare both the accuracy and the average runtime to make a prediction by RED with the baseline approaches, including RAM with the same set of parameters, a 3-layer ConvNet, a 4-layer ConvNet, and RED where the attention $\hat{a}_t$ is uniformly randomly selected from $\cA_t=[-1,1]^2$.
The last baseline is used to show the necessity of the learned attention mechanism.
As shown on Table~\ref{tbl:stained}, RED significantly outperforms all baselines in terms of accuracy, and all attention-based models have better speed compared with ConvNet-based algorithms.

\begin{table}
\centering
\begin{tabular}{lrr}  
\toprule
Approach  & Runtime (s) & Accuracy (test) \\
\midrule
RED         &$0.06$    & $84.43$\%\\
Random $\ca$ &$0.06$    & $51.79$\%\\
RAM         &$0.06$    & $62.35$\%\\
ConvNet-3    &$1.95$       & $81.49$\%\\
ConvNet-4    &$3.30$       & $82.92$\%\\
\bottomrule
\end{tabular}
\caption{Comparisons of RED with different baseline approaches on Stained MNIST.}
\label{tbl:stained}
\end{table}

\subsection{Diabetic Retinopathy Screening}

Diabetic Retinopathy (DR) \cite{fong2004retinopathy} is among the leading causes of blindness in the working-age population of the developed world.
Its consequence of vision loss is effectively prevented by population-wise DR screening, where automatic and efficient DR screening is an interesting problem in medical image analysis.
The screening process is to detect abnormality from the fundus photographs, which are generally in high resolution and are noisy due to the photo-taking procedure.
The high-resolution, low signal-to-noise ratio, and the need for efficient population-wise screening agree with the characterizations of our proposed RED model, which motivates us to test the model on this task.
We test and compare the performance using a dataset publicly available on Kaggle\footnote{https://www.kaggle.com/c/diabetic-retinopathy-detection}. 
While the images are originally rated with five levels, we consider level $0$ and $1$ as negative results $\cY=0$ and level $2$, $3$ and $4$ as positive results $\cY=1$.
The results are shown in Table~\ref{tbl:dr}, where the same hyper-parameters are used as is in the stained MNIST experiment.

\begin{figure}[t]
\includegraphics[width=0.49\textwidth,height=0.68\textwidth]{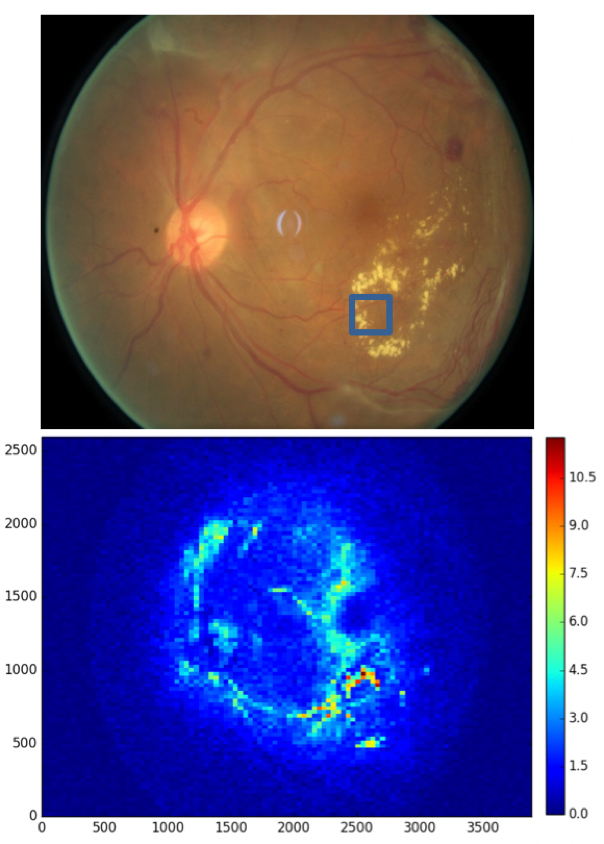}
\caption{Distribution of the attentions in a rollout of RED.}
\label{fig:dr-rollout}
\end{figure}

The performance of our RED approach is compared with RAM and ConvNet with both four layers and five layers.
Also, we test ConvNet with fractional max-pooling layers \cite{graham2014fractional} and cyclic pooling layers \cite{dieleman2015rotation} which have solid performances on the Kaggle challenge.
We re-implement their approach with 4 and 5 layers (\textit{ConvNet-4+} and \textit{ConvNet-5+}) and the comparisons are shown in Table~\ref{tbl:dr}.

Our RED approach achieves extraordinary speed performance while demonstrating competitive accuracy.
Notably, compared with the ConvNet-based methods which usually take many seconds to process each image, RED provides a way to trade marginal accuracy to significant speed improvement.
That could be critical especially for the DR screening tasks designed to be used on population-wise datasets while requiring timely results.
Apart from the speed improvement, it is worth to note that RED is also light weighted, where the number of parameters needed is relatively low to process only small patches at any time step.
The experiments on DR screening demonstrate that our RED method is practical enough to be applied to real-world systems.

\begin{table}
\centering
\begin{tabular}{lrr}  
\toprule
Approach  & Runtime (s) & Accuracy (test) \\
\midrule
RED         &$0.04$    & $91.55$\%\\
Random $\ca$ &$0.04$    & $53.44$\%\\
RAM         &$0.04$    & $81.35$\%\\
ConvNet-4    &$2.32$       & $90.61$\%\\
ConvNet-4+   &$2.32$       & $91.97$\%\\
ConvNet-5    &$2.92$       & $91.84$\%\\
ConvNet-5+   &$2.92$       & $92.29$\%\\
\bottomrule
\end{tabular}
\caption{Comparisons of RED with different baseline approaches on DR screening.}
\label{tbl:dr}
\end{table}

\subsection{Intuitive Demonstration of the Trajectory}

To understand the policy that deploys the agent's attention, we present a graphical demonstration of the trajectory, which imitates the way humans process existence detection tasks.
As shown in Figure~\ref{fig:dr-rollout} top, the trained agent predicts if patterns related to DR exist in a fundus image.
To observe the trajectory, we put the limit $T\rightarrow \infty$ on the time horizon while keeping the stochastic components $\epsilon$ in Equation~\eqref{eqn:att}.
We then illustrate the distribution of the attentions, in the form of a heat map, in Figure~\ref{fig:dr-rollout} bottom.

We first observe that the attention are majority crowded in the bottom right part of the image, which coincides with the lesion patterns (yellow stains on the fundus image).
Within the small blue box marked on Figure~\ref{fig:dr-rollout} top concentrates 30 out of the first 250 saccades.
It shows the ability of the trained model to locate regions of interest and to deploy its limited attention resource selectively.
Notably, only 4 of them happen in the first 100 time steps, and the density of attention for $T\rightarrow \infty$ becomes even higher ($\geq 9$ heat value on Figure~\ref{fig:dr-rollout} bottom).
On the other hand, we observe that the model tends to deploy its attentions on around the blood vessels especially at the early stages of the process.
Such behavior helps the agent to gain information about the context of the image and locate the region of interest in the later stages of the process.
Also, it is worth to note that the agent does not get stuck in a small region even when we set the time horizon to be arbitrage large.
Instead, the agent keeps exploring the image indefinitely.
The way the agent automatically balance exploitation and exploration is what we have been expecting an RL algorithm to learn.

\section{Conclusion and Future Works}

We present recurrent existence determination, a novel RL algorithm for existence detection.
RED imitate the attention mechanism that humans elaborate to process object detection both efficiently and precisely, yielding similar characterizations as desired.
RED employs hard attention which boosts the test-time speed while the non-differentiability introduced by the attention mechanism is addressed via policy optimization.
We propose the $k$-maximum aggregation layer and other components in RED which help to solve the delayed reward problem and automatically learns to balance exploration and exploitation.
Experimental analysis shows significant speed and accuracy improvement compared with previous approaches, on both synthetic and real-world datasets.

One of the plausible future direction is to further address the delayed reward problem, by adding a value network as the critic the actor-critic method \cite{sutton2018reinforcement,mnih2016asynchronous}.
The critic will give the agent immediate feedback for any action it takes, using the estimation of the action-state value function.
In this case, as the environment is partially observable, the actor-critic need to be asymmetric where the critic will have access to the full image.
The critic network is expected to be a proper replacement of the aggregation layer in this paper with an improved performance.

\newpage

\bibliographystyle{named}
\bibliography{ijcai19}

\end{document}